\documentclass[10pt,twocolumn,letterpaper]{article}

\usepackage{cvpr}
\usepackage{times}
\usepackage{epsfig}
\usepackage{graphicx}
\usepackage{amsmath}
\usepackage{amssymb}
\usepackage[table]{colortbl}
\definecolor{bed}{rgb}{0.161, 0.416, 0.42}
\definecolor{books}{rgb}{0.882, 0.235, 0.207}
\definecolor{ceiling}{rgb}{0.267, 0.137,0.741}
\definecolor{chair}{rgb}{0.047,0.22,0.957}
\definecolor{floor}{rgb}{0.781,0.781,0.91}
\definecolor{furniture}{rgb}{0.91,0.278,0.161}
\definecolor{objects}{rgb}{0.898,0.212,0.447}
\definecolor{painting}{rgb}{0.125,0.173,0.506}
\definecolor{sofa}{rgb}{0.812,0.651,0.49}
\definecolor{table}{rgb}{0.427,0.93, 0.165}
\definecolor{tv}{rgb}{0.98,0.8,0.157}
\definecolor{wall}{rgb}{0.518, 0.518, 0.518}
\definecolor{window}{rgb}{0.373,0.973,0.984}

\usepackage[breaklinks=true,bookmarks=false]{hyperref}

\cvprfinalcopy

\def\cvprPaperID{****} % *** Enter the CVPR Paper ID here

\begin{document}

\title{Semi-Dense 3D Semantic Mapping from Monocular SLAM}

\author{Xuanpeng LI\\
Southeast University\\
2 Si Pai Lou, Nanjing, China\\
{\tt\small li\_xuanpeng@seu.edu.cn}
\and
Rachid Belaroussi\\
IFSTTAR, COSYS/LIVIC\\
25 all\'ee des Marronniers, 78000 Versailles, France \\
{\tt\small rachid.belaroussi@ifsttar.fr}
}

\maketitle

\begin{abstract}
The bundle of geometry and appearance in computer vision has proven to be a promising solution for robots across a wide variety of applications. Stereo cameras and RGB-D sensors are widely used to realise fast 3D reconstruction and trajectory tracking in a dense way. However, they lack flexibility of seamless switch between different scaled environments, i.e., indoor and outdoor scenes. In addition, semantic information are still hard to acquire in a 3D mapping. We address this challenge by combining the state-of-art deep learning method and semi-dense Simultaneous Localisation and Mapping (SLAM) based on video stream from a monocular camera.  In our approach, 2D semantic information are transferred to 3D mapping via correspondence between connective \emph{Keyframes} with spatial consistency. There is no need to obtain a semantic segmentation for each frame in a sequence, so that it could achieve a reasonable computation time. We evaluate our method on indoor/outdoor datasets and lead to an improvement in the 2D semantic labelling over baseline single frame predictions.
\end{abstract}

\section{Introduction}
%\subsection{background}
Understanding 3D scene is more widely required but still challenging in many robotics applications.   For instance, autonomous navigation in outdoor scene asks for a comprehensive understanding of immediate surroundings. In domestic robotics, a simple fetching task always requires knowledge of both what something is, as well as where it is located~\cite{mccormac2016semanticfusion}. Semantic segmentation is an important and promising step to address this problem. The state-of-art Convolutional Neural Networks (CNNs) make great advances in the image-based 2D semantic segmentation.  Combined with SLAM technology, mobile robotics could locate itself and meanwhile recognise objects in pixel-wise level.  It means that a task like ``move the chair behind the nearest desk'' or ``park ego-car in front of the left red one at the parking space'' could be accurately accomplished. However, scaled sensors, such as stereo  or RGB-D cameras, only provide reliable measurements in their limited range.  They lack of flexibility of seamless switch between indoor and outdoor scenes. In this work, we exploit a Large-scale Direct Monocular SLAM (LSD-SLAM)~\cite{engel2014lsd} provides cues of 3D spatial information working in both indoor and outdoor scenes and combine with recent advances of DeepLab-based CNN~\cite{DBLP:journals/corr/ChenPK0Y16} to build a 3D scene understanding system. 

%\subsection{problem}
Most man-made environments, no matter indoor or outdoor scenes, usually exhibit distinctive spatial relations amongst varied classes of objects. Being able to capture, model and utilise these kinds of relations could enhance semantic segmentation performance~\cite{wolf2015fast}. In this paper, apart from semi-dense 3D mapping based on monocular SLAM and 2D semantic segmentation via using trained CNN model, 2D-3D transfer and map regularisation  in the framework of semi-dense 3D reconstruction are considered as our main research contribution.  

\begin{figure}
\centering
\includegraphics[width=0.48\textwidth]{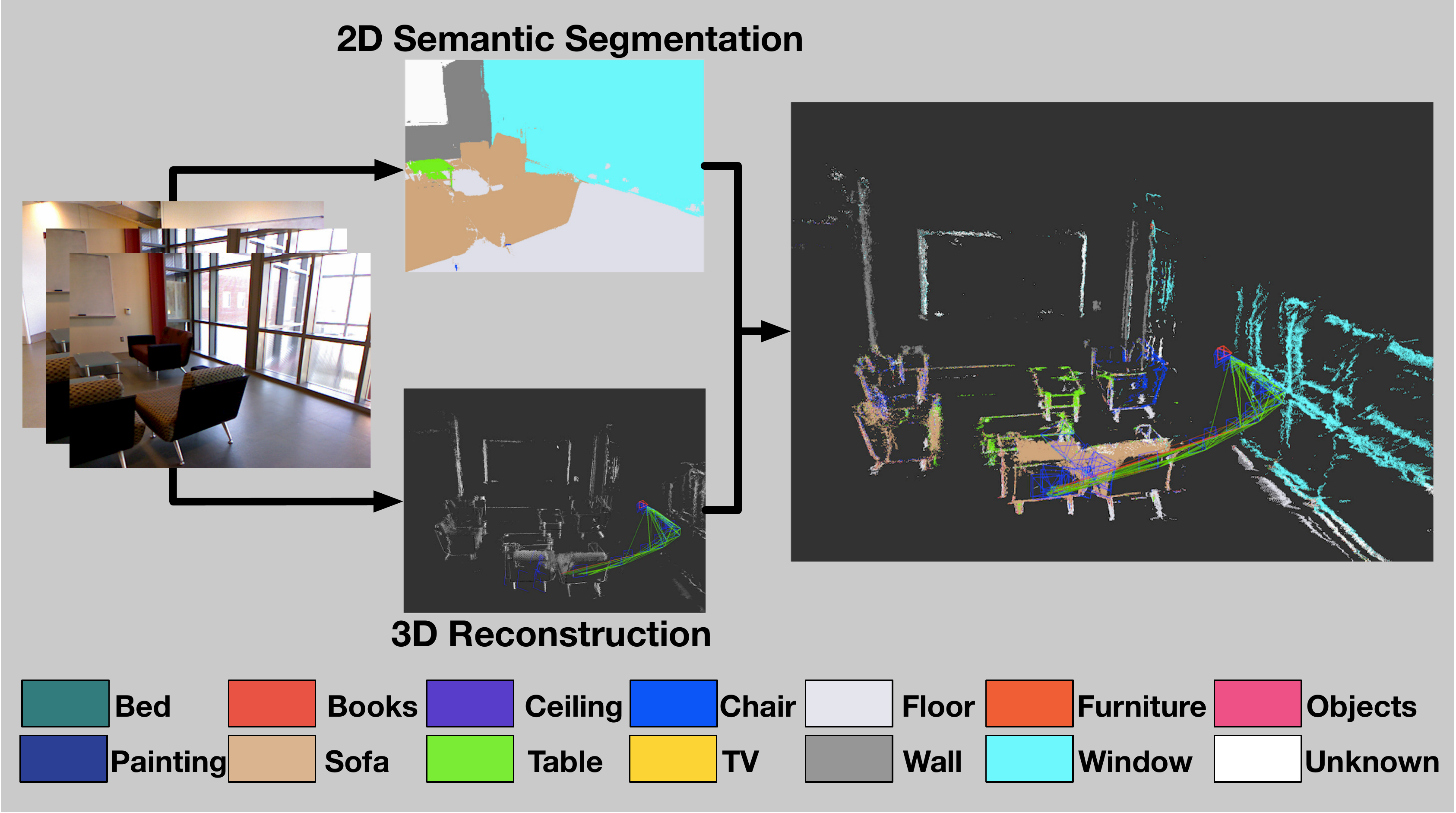}
\caption{\textbf{semi-dense 3D semantic mapping:} The figure shows an example of our system. The sequence of RGB images is used to reconstruct 3D environments and 2D semantic label is predicted on the input frame. Using a 2D-3D label transfer approach and map regularisation could improve the labelling accuracy in a semi-dense way.  }
\label{overall}
\end{figure}

%\subsection{approach }
Our approach is to use stat-of-the-art deep CNN components to predict semantic information which are projected to globally consistent 3D map from a real-time monocular SLAM system. The 3D map is incrementally constructed by a sequence of selected frames with calculated depth information as tracking references. This allows the 2D CNN's semantic labelling attached to keyframes which can be fused into the 3D map in a semi-dense way, as shown in Figure~\ref{overall}.  There is no need to segment each frame in a sequence, which  could save a considerable amount of computation cost. Since the 3D map should have global consistent depth information, it would be regularised in light of its geometry. The regularisation process after the 2D-3D transfer is aimed to remove distinctive outliers and makes the components in the 3D map more consistent, i.e., local points with semantic label should be close in space. NYUv2 and CamVid/KITTI datasets were chosen to evaluate our approach, and we witness an improvement in 2D semantic segmentation. The unlabelled raw videos were used to reconstruct 3D map with semantic predictions in real-time ($\approx$10Hz).

\begin{figure*}
\centering
\includegraphics{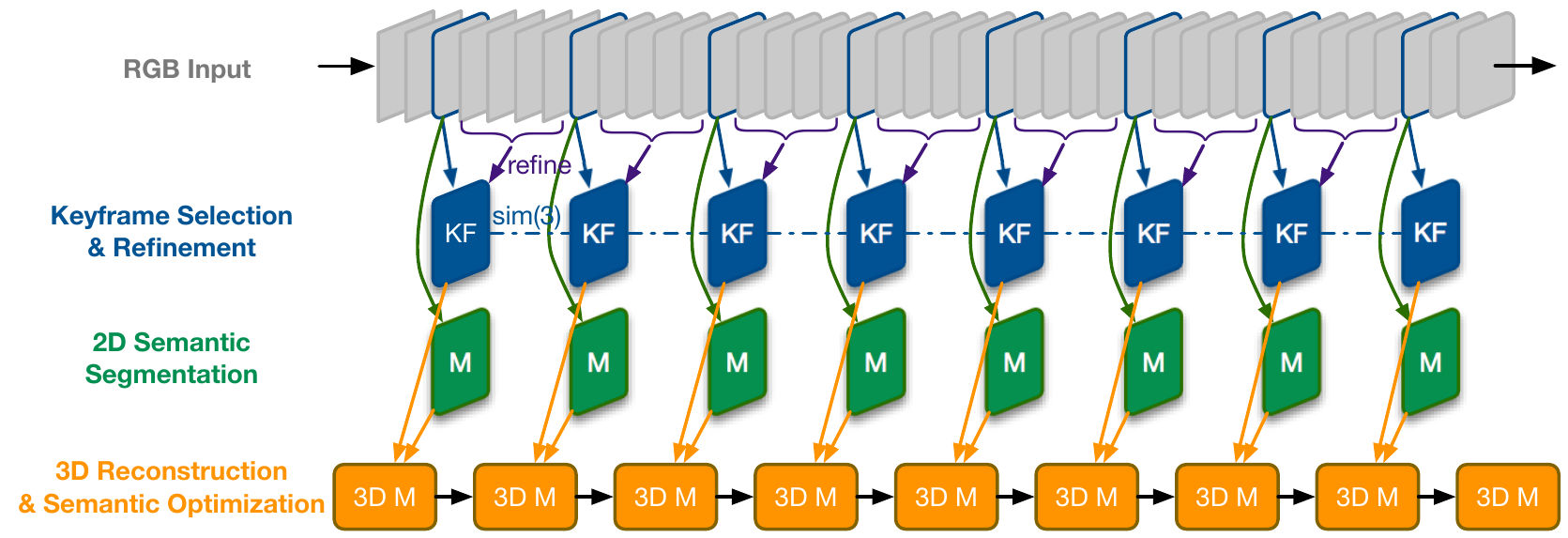}
\caption{\textbf{Overview of our method:} The input is only RGB frame in sequence. There are three separate processes, a Keyframe selection process, a 2D semantic segmentation process , and a 3D reconstruction with semantic optimization process. Keyframes are selected from the sequence of frames as tracking reference and the consecutive frames refine the depth and variance contained in Keyframes. The 2D semantic segmentation process classifies the image information in Keyframes. Finally, the 3D map is  reconstructed with the incoming Keyframes and fused with accumulated semantic information. The 3D semantic map is then regularised  by a dense CRF. The visualised intervals do not correspond to the  actual situation.}
\label{framework}
\end{figure*}

%\subsection{summary}

The paper is presented as follows. The next section gives an overview of related work. Section~\ref{Approach} describes the main components in this work. Section~\ref{Experiments and Results} discusses our experimental results of both indoor and outdoor scenes.

\section{Related Work}
Our work is motivated by~\cite{mccormac2016semanticfusion} which contributes an indoor 3D semantic SLAM from a RGB-D input. They aimed towards a dense 3D map based on ElasticFusion SLAM~\cite{whelan2015elasticfusion} with semantic labelling. Pixel-wise semantic information is acquired from a Deconvolutional Semantic Segmentation network~\cite{Noh_2015_ICCV} with scaled RGB input and Depth as fourth channel.  Depth information is also used to update surfel's depth and normal information to construct 3D dense map during loop closure.  This system required a RGB-D input with reliable depth measurements, but it limited the scene switch. In addition, previous work, SLAM++~\cite{salas2013slam++}, created a map with semantically defined objects, but it is limited to predefined database and hand-crafted template models. 

%\subsection{Visual Slam : Slam++, Dense slam, SemiDense slam, et al.}
Visual SLAM usually contains sparse, semi-dense, and dense types depending on the methods of image alignment. Feature-based methods only exploited limited feature points - typically image corners and blobs or line segments, such as classic MonoSLAM~\cite{davison2007monoslam} and ORB-SLAM~\cite{mur2015orb}. They are not suitable for 3D semantic mapping due to rather limited feature points. In order to better exploit image information and avoid the cost on calculation of features,  direct dense SLAM system, such as surfel-based ElasticFusion~\cite{whelan2015elasticfusion} and Dense Visual SLAM~\cite{kerl2013dense}, have been proposed recently. Whereas, direct image alignment from these dense methods is well-established for RGB-D or stereo sensors~\cite{engel2014lsd}, not for monocular camera.  Semi-dense methods like LSD-SLAM~\cite{engel2014lsd}  and SVO~\cite{forster2014svo} provide possibility to build a synchronised 3D semantic mapping system.

%\subsection{Fcn, DeconvNet, DeepNet, DenseCRF}
Deep CNN has proven to be successful in the field of image semantic segmentation. Long \emph{et al.}~\cite{long2015fully} firstly introduced an inverse convolution layer to realise an end-to-end training and inference process. Then, an encoder-decoder architecture with max unpooling and deconvolutional layers was proposed to avoid the separate step training problem in FCN network~\cite{Noh_2015_ICCV}.  The cutting-edge method, namely, DeepLab,  combines atrous convolutions and atrous spatial pyramid pooling (ASPP) to achieve a state-of-the-art performance on semantic segmentation. It also incorporates dense Conditional Random Field (CRF) which improves both qualitatively and quantitatively via a posterior process. 

%\subsection{ Dense 3D, fast 3D, et al.}
Our semi-dense approach is also inspired by dense 3D semantic mapping methods~\cite{hermans2014dense, wolf2015fast, sengupta2013urban, martinovic20153d} in both indoor and outdoor scenes. The major contributions from these work are 2D-3D transfer and map regularisation.  Especially,  Hermans \emph{et al.} proposed an efficient 3D Conditional Random Fields to regularise consistent 3D semantic mapping considering influence between neighbors of 3D points (voxels).   In this work, we explore a similar strategy of utilising their semantic, visual, and geometrical information to enforce spatial consistency.

\section{Approach}
\label{Approach}
Our target is to create a 3D scene map with semi-dense and consistent label information online while robotics equipped with a monocular camera move through an unknown scene.   The approach is decoupled into three separately running  processes as shown in Figure~\ref{framework}. The 3D reconstruction process selects keyframes from the sequence of image frames captured by monocular camera. The selected keyframes are stacked to reconstruct the 3D map based on their pose graphs. This whole  process runs within CPU in real time. Meanwhile, the 2D semantic segmentation process predicts the pixel-level classification on keyframes. The depth information of keyframes are iteratively  refined  by  their consecutive frames. It creates local optimal depth estimation for each keyframe and correspondence  between labelled pixels and voxels in the 3D point  cloud. To obtain a globally optimal 3D semantic segmentation, we exploit the information over neighboring 3D points, involving the distance, color similarity and semantic label. This process achieves the update of 3D point's state and make a globally consistently 3D map. The following section describes each process in more detail. 

\subsection{2D Semantic Segmentation}
In our work, deep CNN adopts the core layers as the DeepLab-v2 proposed by Chen \emph{et al.}~\cite{DBLP:journals/corr/ChenPK0Y16}. Two important components in DeepLab-v2 are the dilated convolution or named as atrous convolution and atrous spatial pyramid pooling (ASPP), which enlarge the field of view of filters and explicitly combine the feature maps of multiple scales. The final result comes from fusion of the bilinear interpolation of  multi-scaled feature maps to the original image resolution. This method can capture the details  and successfully handle both large and small objects in images. The encoder part of entire architecture is built based on the VGG 16-layer network. For the inference, we use a softmax layer  to obtain the final probabilistic score map. 

For the indoor scene, we employ NYUv2 labelled dataset and adopt  the predefined 14 classes, denoted as $\mathcal{L} = \{l_1, l_2, \dots, l_{\mathrm{M}}\}(\mathrm{M}=14)$,  including floor, wall, sofa, and so on~\cite{couprie2013indoor}. The NYUv2 dataset consists of 795 labelled training images and 654 labelled val images.  On the other hand, we use CamVid dataset for training towards the outdoor scene within 11 classes, including sky, building, car, road and so on. The CamVid dataset is split into 367 labelled training images, 101 labelled val images and 233 labelled test images. Since there is no sequence of images in CamVid dataset, we use all labelled images for training and test our whole approach on the KITTI dataset. We finetune our model  with COCO pre-trained model without depth information involved. We do not rescale our input images to the native resolution $(224\times224)$ as a traditional VGG16 network. The dimension of input in the training process is cropped to $321\times321\times3$ due to the alignment of map  after multiple scales subsampling. During the inference, we keep the original resolution of input image according to different datasets.

\subsection{Semi-Dense SLAM Mapping}
LSD-SLAM is a real-time and semi-dense 3D mapping method. The 3D environment is reconstructed as pose-graph  keyframes with associated semi-dense depth maps. The keyframe is selected from image frames in light of its distance from previous keyframe as tracking reference. Each  keyframe (index $i$) $\mathcal{K}_i = \{I_i, D_i, V_i, S_i \}$ consists of an image intensity $I_i: \Omega_i \rightarrow \mathbb{R}^3$, a depth map $D_i: \Omega_{D_i} \rightarrow \mathbb{R}$, the variance of depth map $V_i: \Omega_{D_i} \rightarrow \mathbb{R}$ and a semantic score map $S_i: \Omega_i \rightarrow \mathbb{R}^3$.  Depth map and variance are defined  in subset of pixels as  $\Omega_{D_i} \subset \Omega_i$, which means semi-dense, only available for certain image regions of large intensity gradient.  The semantic score  map is with a  size of $\mathrm{H}\times\mathrm{W}\times\mathrm{M}$ ($\mathrm{H}$: height, $\mathrm{W}$: width, $\mathrm{M}$: number of classes) directly from deep CNN.

Spatial regularisation and  outlier removal are incorporated in the creation of depth map  and tracked image frames are used to refine depth map based on small-baseline stereo comparisons~\cite{engel2013semi}. Next, directly scale-drift aware image alignment on $\mathfrak{sim}(3)$ is carried on these stacked keyframes with refined depth map, which is used to align two differently scaled keyframes~\cite{engel2014lsd}. The scale-drift aware operation is against different scale environments, such as office rooms (indoor) and urban city road (outdoor). Due to the inherent correlation between depth map and tracking accuracy, depth residual is exploited to estimate the scaled transformation between keyframes. Consequently, we build a 3D point cloud based on the depth maps of keyframes with minimum of error of image alignment.  It could run in real time on a CPU about 25 Hz.

\subsection{Accumulated Labelling Fusion}
Single 2D semantic segmentation would have inconsistent labelling between consecutive frames due to uncertainty of sensors and environments. Incremental fusion of semantic label information of stacked keyframes is similar to SLAM correspondences that allow  us to associate probabilistic label from multiple keyframes in a Bayesian way, like the approach in~\cite{mccormac2016semanticfusion}. For the 3D map at current keyframe $\mathcal{K}_t$, we denote the class distribution of a 3D voxel $\mathbf{v}_d$ as  $l_k$, where our goal is to obtain the each 3D point independent probability  over the class labels $p(\mathbf{v}_d \rightarrow l_k \vert \mathcal{K}_0^t)$ given all stacked keyframes $\mathcal{K}_0^t = \{\mathcal{K}_0, \mathcal{K}_1, \dots, \mathcal{K}_t\}$. We use a recursive Bayes' rule to transfer this:
\begin{equation}
p(\mathbf{v}_d\rightarrow l_k \vert \mathcal{K}_0^t) = \frac{1}{Z_i}p(\mathcal{K}_t\vert\mathcal{K}_0^{t-1}, l_k)p(l_k\vert\mathcal{K}_0^{t-1}),
\end{equation}
where $Z_t = p(\mathcal{K}_t\vert\mathcal{K}_0^{t-1})$. Applying first Markov assumption to $p(\mathcal{K}_t\vert\mathcal{K}_0^{t-1}, l_k)$, then we  have:
\begin{eqnarray}
p(\mathbf{v}_d\rightarrow l_k \vert \mathcal{K}_0^t)  &=& \frac{1}{Z_t}p(\mathcal{K}_t\vert l_k)p(l_k\vert\mathcal{K}_0^{t-1}) \\
&=& \frac{1}{Z_t} \frac{p(\mathcal{K}_t)p(l_k\vert\mathcal{K}_t)}{p(l_k)}p(l_k\vert\mathcal{K}_0^{t-1}). \nonumber
\end{eqnarray}
We assume $p(l_k)$ does not change over time and there is no need to calculate the normalisation factor $p(\mathcal{K}_t)/Z_t$ explicitly. Finally, we can update the semantic information of 3D point cloud when a new keyframe arrives as follows:
\begin{equation}
p(\mathbf{v}_d\rightarrow l_k \vert \mathcal{K}_0^t) \thicksim p(l_k\vert\mathcal{K}_t)p(l_k\vert\mathcal{K}_0^{t-1}).
\end{equation}
This incremental fusion of semantic probabilistic information allows us to label 3D point based on whole existing keyframes in real-time. The following section describes how we explore dense CRF to regularise semi-dense 3D map using map geometry, which could propagate semantic information between spatial neighbors. 

\subsection{Semi-Dense Map Regularisation}
Dense CRF is widely employed in 2D semantic segmentation to smooth noisy segmentation map. Some previous works~\cite{sengupta2013urban, hermans2014dense, wolf2015fast} seek its application on 3D map to model contextual relations between various class labels in a fully connected graph. This algorithm aims at minimising the Gibbs energy $E$ by means of mean-field approximation and message passing scheme to efficiently infer the latent variables. In a 3D environment, a voxel $\mathbf{v}_d$ in the point cloud $\mathbf{P}$ is assigned a label $l\in\mathcal{L}$. Then, a whole label assignment $\mathbf{l}\in \mathcal{L}^M$ has a corresponding Gibbs energy that consists of unary and pairwise potentials $\psi_i$ and $\psi_{i,j}$:
\begin{equation}
E(\mathbf{l}\vert\mathbf{P, \theta}) = \sum_i\psi_i(l_i\vert\mathbf{P}, \theta) + \sum_{i<j}\psi_{i,j}(l_i, l_j\vert\mathbf{P}, \theta)
\end{equation}
with $1 \le i, j \le M$.

The unary potential is defined as the negative logarithm of the labelling's probability:
\begin{equation}
\psi_i(l_i\vert\mathbf{P},\theta) = -\log(p(\mathbf{v} \rightarrow \l_i\vert\mathcal{K}_0^t)).
\end{equation}
The pairwise potential is modeled to be a linear combination of $m$ Gaussian edge potential kernels:
\begin{equation}
\psi_{i,j}(l_i, l_j\vert\mathbf{P},\theta) = \sum_m\mu^{(m)}(l_i, l_j\vert \theta)k^{(m)}(\mathbf{f}_i, \mathbf{f}_j),
\end{equation}
where $\mu^{(m)}$ is a label compatibility function corresponding to the kernel functions $k^{(m)}(\mathbf{f}_i, \mathbf{f}_j)$ and $\mathbf{f}$ denotes feature vector for voxel, $v$.

For our application in 3D environments, we explore two Gaussian kernels for the pairwise potentials, similar to the work of Hermans \emph{et al.}~\cite{hermans2014dense}. The first one is a spatial smoothness kernel as Eq.~\ref{smoothpotential}, which aims at  enforcing a local, appearance-agnositc smoothness amongst voxels with similar normal vectors. 
\begin{equation}
\label{smoothpotential}
k^{(1)}=\omega^1 \mathrm{exp}\Bigg( -\frac{\vert \mathbf{p}_i -\mathbf{p}_j\vert^2}{2\theta^2_{p,n}} - \frac{\vert\mathbf{n}_i - \mathbf{n}_j\vert^2}{2\theta^2_{n}} \Bigg),
\end{equation}
where $\mathbf{p}$ are the coordinates of  3D voxels and $\mathbf{n}$ are the respective surface normals. 

Most researches employ  an appearance kernel as the second one,
\begin{equation}
k=\omega \mathrm{exp}\Bigg( -\frac{\vert \mathbf{p}_i -\mathbf{p}_j\vert^2}{2\theta^2_{p,c}} - \frac{\vert\mathbf{c}_i - \mathbf{c}_j\vert^2}{2\theta^2_{c}} \Bigg),
\end{equation}
where $\mathbf{c}$ are the RGB/LAB color vectors of the corresponding voxels~\cite{mccormac2016semanticfusion}. However, semi-dense LSD-SLAM only utilises the points of high intensity gradient to reconstruct 3D environment. These points are rather limited to capture contextual relations between different classes.  

Thus, we take a semantic score-related kernel in Eq.~\ref{semanticpotential}, which encourages the voxels in a given segment to take the same label and penalises partial inconsistency of voxels as similar as the work in~\cite{sengupta2013urban}. 
\begin{equation}
\label{semanticpotential}
k^{(2)} = \omega^2 \mathrm{exp}\Bigg( -\frac{\vert \mathbf{p}_i -\mathbf{p}_j\vert^2}{2\theta^2_{p,s}} - \frac{\vert\mathbf{s}_i - \mathbf{s}_j\vert^2}{2\theta^2_{s}}\Bigg),
\end{equation}
where $\theta_{p,s} \gg \theta_{p,n}$ means we hope that semantic information flows across larger distances than geometrical structure and $\mathbf{s}$ is the probabilistic score of all classes.

In addition, we take a similar strategy as~\cite{wolf2015fast}, by defining  separate label compatibility functions $\mu^{(m)}$ for both kernels. For the smoothness kernel, we use a simple Potts model: $\mu^{(1)}(l_i, l_j\vert\theta) = 1_{[l_i \neq l_j]}$, while a more expressive kernel is defined for semantic potential to distribute the probabilistic score across different distances. Then,  $\mu^{(2)}$ is a full, symmetric $14 \times 14$ matrix, where all class relations are defined individually. We did not tune these parameters above.  All implementations follow the default settings presented in the work~\cite{koltun2011efficient}. 

\section{Experiments and Results}
\label{Experiments and Results}
\subsection{Experiments on Indoor Scene}
To demonstrate the performance of our approach, we firstly use the publicly available dataset NYUv2 towards the 2D semantic segmentation. We find that the ``poly'' stochastic gradient descent is better than the ``step''  one with a learning rate of $0.001$, a step size of $2000$,  momentum of $0.9$, and weight decay of $5\times10^{-4}$. For each iteration, we use 10 batch for training and the number of  total  iterations is $10k$, which runs on a Nvidia Titan X GPU about 6 hours. The results of our evaluation are presented in Table~\ref{2Dsemantictable}.
%and qualitative results are exampled in Figure~\ref{2Dsemanticfigure}. 
We find that the results of 2D semantic segmentation on the NYU v2 dataset gain an improvement over previous work listed in~\cite{mccormac2016semanticfusion}. Especially, it should be noted that RGBD~\cite{mccormac2016semanticfusion} and Eigen~\cite{eigen2015predicting} use depth information during the training and inference, while we only use the RGB image. In our work, the posterior dense CRF of 2D semantic segmentation contributes an improvement of 1.8\% on the deep CNN on the value of pixel average, but it influence the real-time capability heavily. Thus, for online, it is disabled entirely and we apply it only on the final 3D map regularisation. 

\begin{table*}[]
\centering
\begin{tabular}{|l|p{.5cm}<{\centering}|p{.5cm}<{\centering}|p{.5cm}<{\centering}|cp{.5cm}<{\centering}|p{.5cm}<{\centering}|p{.5cm}<{\centering}|p{.5cm}<{\centering}|p{.5cm}<{\centering}|p{.5cm}<{\centering}|p{.5cm}<{\centering}|p{.5cm}<{\centering}|p{.5cm}<{\centering}|p{.5cm}<{\centering}|p{.5cm}<{\centering}|p{.5cm}<{\centering}|}
\hline
Method & \cellcolor{bed}\rotatebox{90}{\color{white}bed} & \cellcolor{books}\rotatebox{90}{books} & \cellcolor{ceiling}\rotatebox{90}{\color{white}ceiling} & \cellcolor{chair}\rotatebox{90}{\color{white}chair}  & \cellcolor{floor}\rotatebox{90}{floor} & \cellcolor{furniture}\rotatebox{90}{furniture} & \cellcolor{objects}\rotatebox{90}{objects} & \cellcolor{painting}\rotatebox{90}{\color{white}painting} & \cellcolor{sofa}\rotatebox{90}{sofa} & \cellcolor{table}\rotatebox{90}{table} & \cellcolor{tv}\rotatebox{90}{tv} & \cellcolor{wall}\rotatebox{90}{wall} & \cellcolor{window}\rotatebox{90}{window} & \rotatebox{90}{class avg.} & \rotatebox{90}{pixel avg.} \\
\hline
Hermans \emph{et al.}~\cite{hermans2014dense} & \textbf{68.4}& 45.4& \textbf{83.4}& 41.9& 91.5& 37.1& 8.6& 35.8& 28.5& 27.7& 38.4& 71.8& 46.1& 48.0 & 54.3 \\
\hline
RGBD-SF~\cite{mccormac2016semanticfusion} &61.7 &\textbf{58.5} &43.4 &58.4 &92.6 &63.7 &\textbf{59.1} &66.4 &47.3 &34.0 &33.9 &86.0 &60.5 &58.9 &67.5\\
RGBD-SF-CRF &62.0 &58.4 & 43.3 &59.5 &\textbf{92.7} &64.4 &58.3 &65.8 &48.7 &34.3 &34.3 &86.3 &\textbf{62.3} &59.2 &67.9\\
\hline
Eigen-SF~\cite{eigen2015predicting} & 47.8& 50.8& 79.0& 73.3& 90.5& 62.8& 46.7& 64.5& 45.8& \textbf{46.0}& 70.7& 88.5& 55.2& 63.2& 69.3 \\
Eigen-SF-CRF & 48.3& 51.5& 79.0& \textbf{74.7}& 90.8& 63.5& 46.9& 63.6& 46.5& 45.9& \textbf{71.5}& \textbf{89.4}& 55.6& \textbf{63.6}& 69.9\\
\hline
Ours          &62.8 &37.5 &72.0 &64.7 &89.3 &62.4 &19.7 &67.3 &56.2 &41.6 &58.9 &83.7 &54.8 &59.4  &68.5\\ 
Ours-CRF  &64.9 &34.6 &72.0 &67.5 &90.5 &\textbf{65.0} &17.2 &\textbf{67.3} &\textbf{59.3} &41.3 &60.0 &85.1 &57.0 &60.3 &\textbf{70.3}\\
\hline
\end{tabular}
\caption{NYUv2 test set results: It should be mentioned that in the NYUv2 dataset, there are more than 800 classes. The objects here are defined as several stuffs, such as box, bottle and cup, which may be not identical to other's work. It leads to a low score compared to other classes in our evaluation.  Besides, our model is evaluated only based on the training of RGB  input in its original resolution ($640\times480$).}
\label{2Dsemantictable}
\end{table*}

A typical example of indoor 3D environment is shown in Figure~\ref{overall}, which contains about 370k 3D points within a total of 30 generated keyframes (about 12k points for each keyframe). Since our approach only executes the 2D semantic segmentation on keyframes, it could run in 10Hz online. 
We're aware that the initialisation of LSD-SLAM is quite significant to calculate the accurate depth of keyframes in sequence.  Excessive rotation and lacking translation at the beginning would lead to a poor mapping result. We select the sequences in NYU v2 which have effective initialisation. All tests were performed on an Intel Core i7-5930K CPU and a NVIDIA Titan X GPU.

\subsection{Experiments on Outdoor Scene}
\begin{figure*}[!]
\centering
\includegraphics[width=\textwidth]{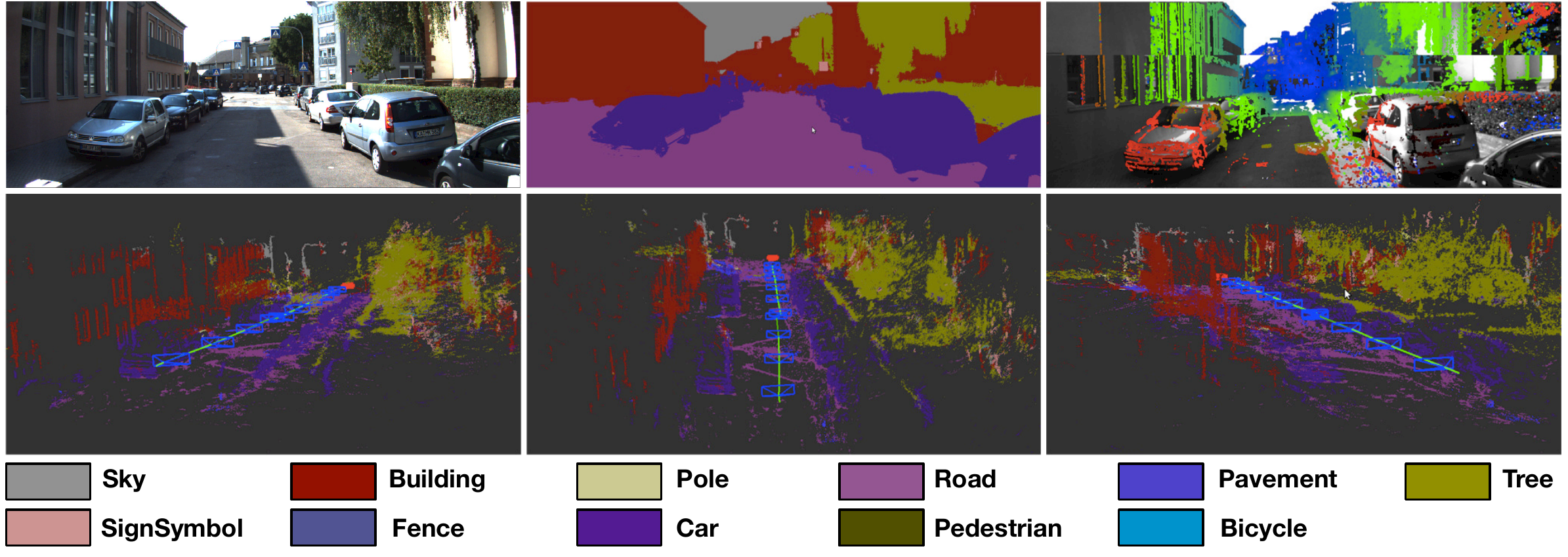}
\caption{\textbf{Qualitative KITTI test set results}. (\textbf{Top Left}) Original image. (\textbf{Top Center}) 2D semantic segmentation. (\textbf{Top Right}) Current keyFrame. (\textbf{Bottom Row}) 3D semantic mapping of a same section from different views.  }
\label{experiment2}
\end{figure*}

Outdoor scenes usually demand larger range of measurement than indoor scenes. 
The training process on the CamVid dataset is similar to NYUv2 dataset, using a ``poly'' strategy within 10 batch in 10k iterations.  Since LSD-SLAM works only based on consecutive images, we need to use image sequence to test our system. Thus, we choose raw image sequences of KITTI dataset  in our experiments.  The resolution of images in KITTI dataset is $1392\times512$. Take the \emph{2011\_09\_26\_drive\_0093} clip as an example in Figure~\ref{experiment2}. There are 439 raw images of this clip with  43 seconds. Our system generates about 1600k 3D points within 56 keyframes (about 30k points for each keyframe). The inference process of KITTI image costs about 400 ms to generate a labelling score map of same resolution. Due to 3D reconstruction of selected keyframes, the 3D mapping could arrive at a speed of 10Hz. Moreover, we remove several keyframes at the beginning of sequence, due to limited initialisation of LSD-SLAM based on KITTI data. 

\section{Conclusions}
We have presented a semi-dense 3D semantic mapping based on a monocular SLAM, which runs on a CPU coupled with a GPU. In contrast to previous work, this system does not use any scaled sensors but a single camera, and only selected frames for 2D semantic segmentation, which reduces the computation time. In addition, our scale-drift aware system could attain seamlessly switch between both indoor and outdoor scenes without extra effort on various scales. We explored a state-of-the-art deep CNN to accurately segment objects in various scenes. Direct monocular SLAM reconstructs a 3D semi-dense environment without any prior depth information, which is suitable for mobile robots working in both indoor and outdoor. The semantic annotations, even with inaccurate labels, are transferred into the 3D map and regularised with a CRF process. It achieves a promising result, and fit for online use. Significantly, we find the geometry of reconstructed map could help to segment objects in different depths. 

In future work, we plan to introduce several state-of-the-art SLAM technologies to improve the initialisation. Research on how labelling boosts 3D reconstruction of SLAM would be an interesting direction. And the deep learning method to solve depth estimation and spatial transformation for SLAM like~\cite{eigen2015predicting} would be another interesting topic. 
{\small
\bibliographystyle{ieee}
\bibliography{ref}
}

\end{document}